\documentclass{article}

\usepackage{neurips_2024}
\usepackage{amsmath}
\usepackage{pifont}
\usepackage{graphicx} 
\usepackage{subcaption}

\usepackage[utf8]{inputenc} %
\usepackage[T1]{fontenc}    %
\usepackage{hyperref}       %
\hypersetup{
 colorlinks=true,
 linkcolor=red,
 citecolor=cyan,
 filecolor=magenta, 
 urlcolor=magenta,
 }
\usepackage{url}            %
\usepackage{booktabs}       %
\usepackage{amsfonts}       %
\usepackage{nicefrac}       %
\usepackage{microtype}      %
\usepackage{xcolor}         %

\title{Matrix-3D: Omnidirectional Explorable  3D World Generation}

\author{
Zhongqi Yang  $^{1}$\footnotemark[1] \quad Wenhang Ge $^{2}$\footnotemark[1] \quad Yuqi Li $^{1,3}$\footnotemark[1]\thanks{These three authors contributed equally to this work.} \quad Jiaqi Chen $^{1}$\footnotemark[2] \quad Haoyuan Li $^{1}$\footnotemark[2]\thanks{These two authors contributed equally to this work.}\\
\textbf{Mengyin An}$^{1}$ \quad \textbf{Fei Kang}$^{1}$ \quad\textbf{Hua Xue}$^{1}$ \quad \textbf{Baixin Xu}$^{1}$ \quad \textbf{Yuyang Yin}$^{1}$ \\  \textbf{Eric Li}$^{1}$ \quad \textbf{Yang Liu}$^{1}$ \quad\textbf{Yikai Wang}$^{4}$ \quad\textbf{Hao-Xiang Guo}$^{1}$\footnotemark[3]\thanks{Corresponding author and project lead.} \quad \textbf{Yahui Zhou}$^{1}$\\\\
$^1$ Skywork AI\\ $^2$ Hong Kong University of Science and Technology (Guangzhou)\\ $^3$ Institute of Computing Technology, Chinese Academy of Sciences \\ $^4$ School of Artificial Intelligence, Beijing Normal University
 \\\\
Project Page: \url{https://matrix-3d.github.io}
}

\begin{document}

\maketitle
\begin{figure}[h]
    \centering
    \includegraphics[width=\linewidth]{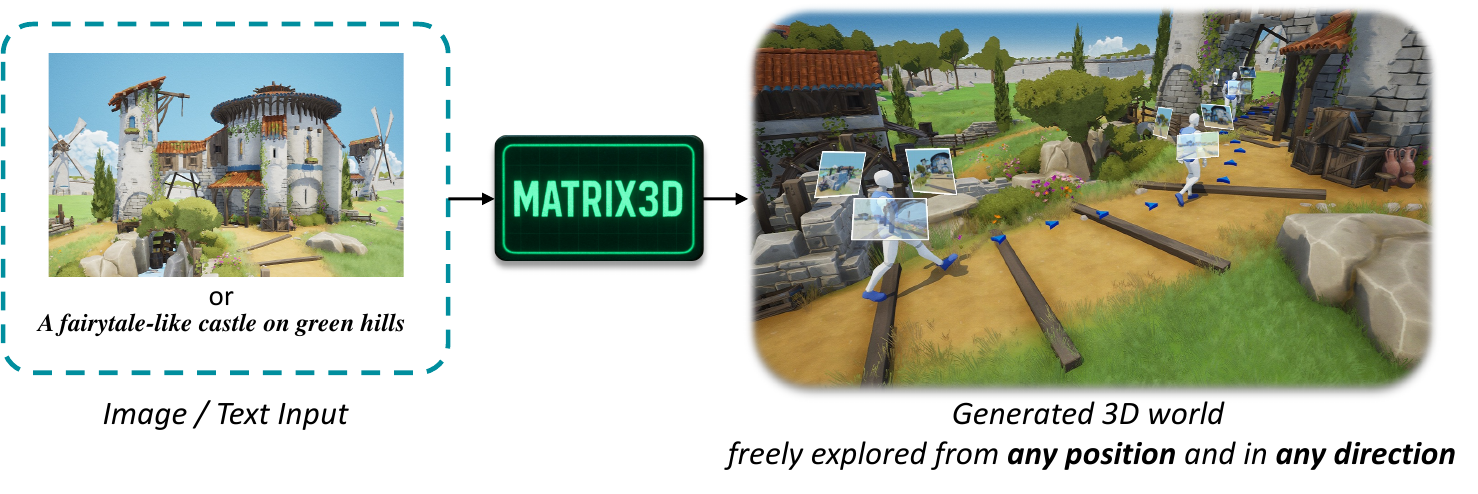}
    \caption{Matrix-3D can generate omnidirectional explorable 3D world from image or text input.}
    \label{fig:teaser}
\end{figure}

\begin{abstract}
 Explorable 3D world generation from a single image or text prompt forms a cornerstone of spatial intelligence. 
 Recent works utilize video model to achieve wide-scope and generalizable 3D world generation.  However, existing approaches often suffer from a limited scope in the generated scenes. 
 In this work, we propose \textbf{Matrix-3D}, a framework that utilize panoramic representation for wide-coverage omnidirectional explorable 3D world generation that combines conditional video generation and panoramic 3D reconstruction.
 We first train a trajectory-guided panoramic video diffusion model that employs scene mesh renders as condition, to enable high-quality and geometrically consistent scene video generation.
 To lift the panorama scene video to 3D world, we propose two separate methods: (1) a feed-forward large panorama reconstruction model for rapid 3D scene reconstruction and (2) an optimization-based pipeline for accurate and detailed 3D scene reconstruction.
To facilitate effective training, we also introduce the \textbf{Matrix-Pano} dataset, the first large-scale synthetic collection comprising 116K high-quality static panoramic video sequences with depth and trajectory annotations.
 Extensive experiments demonstrate that our proposed framework achieves state-of-the-art performance in panoramic video generation and 3D world generation. See more in \url{https://matrix-3d.github.io}.
\end{abstract}

\section{Introduction}

Spatial intelligence is a rapidly developing technology that enables machines to model the 3D world and to understand, analyze, and reason about it. 
The modeling and generation of 3D world are key modules of spatial intelligence.
High-quality, diverse 3D scenes can serve as virtual environments for AI training and testing, enhancing AI's generalization and adaptability in the field of autonomous driving \cite{yurtsever2020survey} and embodied intelligence \cite{duan2022survey}, thus advancing the development of world models. Additionally, 3D scene generation has widespread applications in game design \cite{lewis2002game}, film production \cite{honthaner2013complete}, and virtual reality \cite{schuemie2001research}.

To address the scarcity of 3D scene data and achieve generalizable 3D world generation , 
recent works \cite{yu2024viewcrafter, he2024cameractrl, gao2024cat3d, cao2025uni3c, ren2025gen3c, liang2025wonderland} have explored leveraging diffusion priors for novel view synthesis from a single image, conditioned on the corresponding camera pose. Some approaches \cite{bahmani2024vd3d, bahmani2024ac3d, he2024cameractrl, liang2025wonderland} employ Plücker embeddings as conditioning inputs, while others \cite{yu2024viewcrafter, ren2025gen3c, cao2025uni3c} utilize  point cloud renders, which offer improved camera controllability. However, these methods are typically constrained to narrow-scope generation and reconstruction, as they operate on perspective images that cover only a limited field of view within the 3D world. Consequently, the generated 3D scene is restricted to narrow viewing angles, exposing sharp, unsightly boundary artifacts from other perspectives, which significantly detracts from the immersive experience and limits its utility in downstream applications.

We overcome this limitation by employing panorama images as our intermediate scene representation, as they capture a complete $360^{\circ} \times 180^{\circ}$ view of a scene, enabling a wide-coverage 3D world reconstruction and omnidirectional exploration. A 3D world can then be represented as a sequence of panorama images whose camera centers follow a certain trajectory. In light of this, we design a 3D world generation framework by combining a trajectory guided panoramic video generation model and a follow-up panoramic 3D reconstruction module.

To achieve high-quality panoramic video generation with precise adherence to specified camera trajectories, we choose Wan 2.1 \cite{wan2025wan} as our base model and incorporate scene renders with masks as conditioning inputs. 
Prior works \cite{yu2024viewcrafter, cao2025uni3c, ren2025gen3c, yu2025trajectorycrafter} typically employ point cloud renders as trajectory guidance. However, we notice that such representations often suffer from Moiré artifacts and erroneous occlusion handling, leading to inconsistent geometry perception and degraded visual quality. Instead, we introduce a scene mesh reconstruction method that leverages the initial panorama and its depth, while carefully handling occluded areas to maintain geometric consistency. The scene mesh is then rendered as trajectory condition which guides the conditioned video diffusion model to generate consistent and high-quality videos.

After training the panoramic video generation model, we need to lift the generated 2D content into an omnidirectional explorable  3D world. To this end, we propose two separate reconstruction pipelines.
The first approach is an intuitive optimization pipeline that involves selecting keyframes from the generated panorama video, cropping them to several perspective images and then performing a 3D Gaussian Splatting (3DGS) optimization~\cite{kerbl20233d,fei20243d,lu2024poison,zhang2024mega}. 
In this way, high-quality and detailed 3D scene can be generated. To further improve the speed of 3D scene generation, we propose our second pipeline.
Inspired by recent advances in feed-forward 3D scene reconstruction \cite{liang2025wonderland, ziwen2024long}, we directly reconstruct the 3D world from the latent of the generated panoramic videos for more efficiency.
Specifically, our feed-forward  large panorama reconstruction model takes video latents and corresponding camera trajectory as input, employing a Transformer-based architecture to predict 3D GS attributes, which enable novel view synthesis and support photometric loss computation.
To facilitate effective convergence, We  introduce a two-stage training strategy by first training for panorama depth prediction, followed by panorama Gaussian prediction.

We notice that existing panoramic video datasets \cite{wang2024360dvd, xia2025panowan} primarily contain video-text pairs and lack precise camera pose and geometric annotations, 
which hinders researches on accurate generation and reconstruction. 
Therefore, we further introduce the Matrix-Pano Dataset, a scalable synthetic dataset curated using physics engines such as Unreal Engine 5. It contains 116,759 high-quality static panoramic video sequences, each accompanied by corresponding 3D exploration trajectories, depth maps, and text annotations. 

Extensive experiments validate the effectiveness and superiority of our proposed method. In panoramic video generation, it outperforms existing approaches in both visual quality and camera controllability. For 3D world reconstruction, we design two separate pipelines that  ensure both high efficiency and high-quality results, enabling effective and scalable omnidirectional 3D world reconstruction.

To summarize, our contributions are listed as follows.
\begin{itemize}
    \item We contribute the Matrix-Pano Dataset, a scalable and high-quality panoramic video dataset with precise camera poses, depth maps, and text annotations, tailored for trajectory-guided panoramic video generation and wide-coverage 3D world reconstruction.
    \item We propose a novel trajectory-guided  video diffusion model by scene mesh renders, which effectively alleviates Moiré patterns and incorrect occlusion relationships, leading to improved visual quality in generation.
    \item We propose two types of panoramic 3D reconstruction methods to achieve rapid and detailed 3D reconstruction, respectively.
\end{itemize}

\section{Related Work}

\subsection{3D World Generative Models}
Remarkable progress has been made in object-level 3D generative modeling \cite{hong2023lrm, GS-LRM, ge2024prm, li2025triposg, zhang2024clay, jiang2025dimer}, largely enabled by the emergence of large-scale 3D object datasets. In contrast, 3D world generation remains relatively underexplored, with fewer benchmarks and greater complexity in spatial reasoning and layout.  Recent works \cite{yu2024viewcrafter, ren2025gen3c, gao2024cat3d, sun2024dimensionx, voleti2024sv3d, sargent2024zeronvs} have explored the use of video diffusion models \cite{wan2025wan, kong2024hunyuanvideo, blattmann2023stable} for 3D world generation. These approaches typically follow a two-stage optimization pipeline: In the first stage, the diffusion model generates novel views conditioned on sparse or single-view inputs and target camera poses; in the second stage, per-scene optimization is performed based on the generated views and their associated poses.
Despite their success, such two-stage approaches suffer from efficiency limitations. Recent works \cite{ziwen2024long, liang2025wonderland} explore feed-forward reconstruction models to enable efficient reconstruction of wide-coverage 3D scenes. However, these methods primarily focus on perspective image inputs, which limits their ability to recover omnidirectional 3D structures. 
In this work, we aim to develop an effective feed-forward model and an optimization-based method for panoramic 3D world reconstruction, enabling full-scene recovery. 

\subsection{Camera-controlled Video Diffusion Models}
With the rapid progress of video diffusion models~\cite{sohl2015deep,ho2020denoising,songscore,openai2024worldsim,wan2025wan,zhang2023hipa,kong2024hunyuanvideo,zheng2024memo}, controlling camera motion during video generation has become an increasingly important research focus. Recent works such as MotionCtrl \cite{wang2024motionctrl}, CameraCtrl \cite{he2024cameractrl}, ViewCrafter \cite{yu2024viewcrafter}, TrajectoryCrafter~\cite{yu2025trajectorycrafter} and Matrix-Game~\cite{zhang2025matrixgame} have proposed various strategies to incorporate camera-related information into pretrained video generators. These methods utilize different forms of camera conditioning, including camera extrinsic parameters, Plücker embeddings \cite{sitzmann2021light}, and point cloud renders.
Despite recent advancements, conditioning on extrinsic parameters or Plücker embeddings often fails to provide precise camera control. In contrast, point cloud renders offer improved controllability, but they frequently suffer from Moiré patterns and incorrect occlusion relationships between foreground and background, ultimately degrading generation quality.
In this work, we propose using scene mesh renders with accurate occlusion relationships as a conditioning signal, which leads to improved visual quality of generated videos.
Furthermore, most existing works \cite{ren2025gen3c, cao2025uni3c, bahmani2024ac3d, bahmani2024vd3d, wang2024motionctrl, gao2024cat3d, he2024cameractrl, yu2024viewcrafter} primarily focus on perspective video generation, which limits their applicability to wide-coverage 3D scene reconstruction.

\begin{figure}
    \centering
\includegraphics[width=0.95\linewidth]{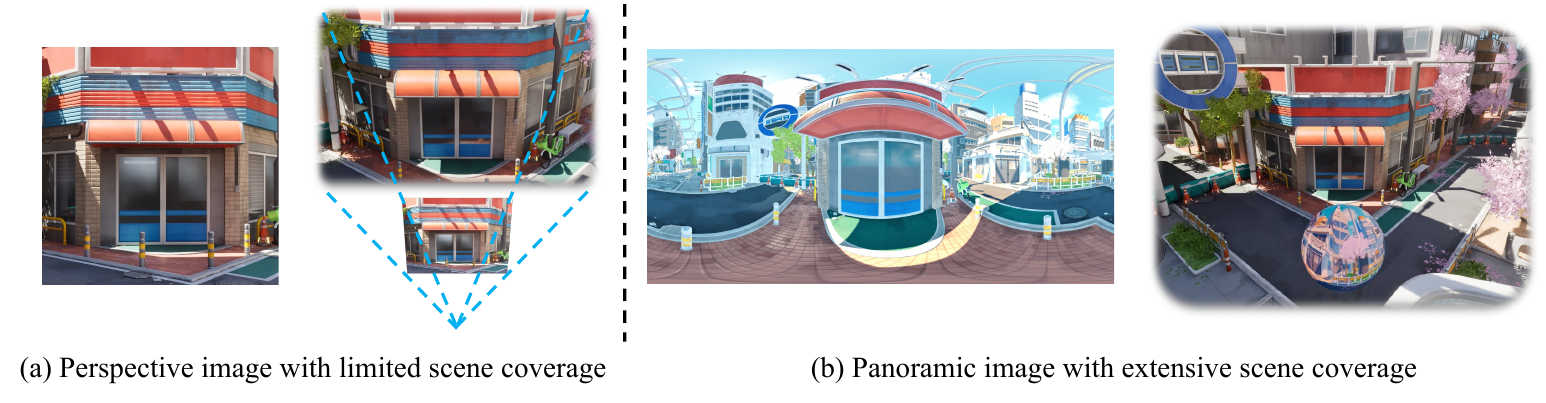}
    \caption{Comparison between perspective and panoramic images. Panoramic images can capture a significantly wider field of view than perspective images.}
    \label{fig:per_pano}
\end{figure}

\subsection{Panorama Generation Models}
Building on recent advances in 2D image synthesis, diffusion-based panoramic image generation models \cite{zhang2024taming, li2023panogen, yang2024layerpano3d, feng2023diffusion360, ye2024diffpano} have achieved notable progress. However, despite offering a full 360-degree field of view, panoramic images inherently lack information about physically occluded regions. To overcome this limitation and capture richer spatial context, panoramic video is required to support the construction of a more complete and spatially coherent world model.
360DVD \cite{wang2024360dvd} introduced the WEB360 dataset, establishing an early benchmark for text-to-panoramic video generation using vision-language models for annotation. More recently, PanoWan \cite{xia2025panowan} presented the PANOVID dataset by aggregating panoramic videos from 360-1M \cite{wallingford2024image}, 360+x~\cite{chen2024360+}, Imagine360 \cite{tan2024imagine360}, WEB360 \cite{wang2024360dvd}, Panonut360 \cite{xu2024panonut360}, and a public immersive VR video dataset \cite{li2017public}, with text descriptions generated by Qwen-2.5-VL \cite{bai2025qwen2}. However, existing datasets provide only video and text prompts, while omitting camera trajectory and geometric information—elements that are essential for constructing 3D-consistent world models. 
Building on these datasets, methods such 4K4DGen~\cite{li20244k4dgen} and DynamicScaler \cite{liu2025dynamicscaler} have advanced dynamic panoramic video generation by modeling realistic object motion. GenEx~\cite{lu2024genex} introduces an action-conditioned panorama generation pipeline. However, none of these methods support the generation of panorama videos with precise scene trajectory control or the conversion of panorama videos into 3D worlds.

\section{Preliminaries}
\label{Preliminaries}
\noindent\textbf{Camera-conditioned Video Diffusion Models.} 
These types of video generation models aim to learn the conditional distribution \( p(z \mid c, s) \), where \( z \) represents the clean video latent produced by a video VAE \cite{wan2025wan, kong2024hunyuanvideo}, \( c \) denotes the conditioning signal (e.g., text or image), and \( s \) corresponds to the camera parameters.
In the flow matching framework \cite{wan2025wan}, the training procedure starts by projecting a clear video $x$ to a latent code $z_1$ with the VAE. Then a noise $z_0 \sim \mathcal{N}(0,I)$ is sampled, and an interpolating latent code $z_t = tz_1 + (1-t)z_0$ is constructed for a sampled timestep $t\in [0,1]$. The training objective is to predict the ground truth velocity $v_t = \mathrm{d}z_t/\mathrm{d}t = z_1 - z_0$, by minimizing the loss function:
\begin{equation}
L(\theta) = \mathbb{E}_{z_0, z_1, c, s, t} \left[ \left\| u_\theta(z_t, c, s, t) - v_t \right\|_2^2 \right],
\end{equation}
where  $u_\theta(z_t, c, s, t)$ denotes the predicted velocity by the model. Following prior methods \cite{yu2024viewcrafter, yu2025trajectorycrafter, ren2025gen3c, cao2025uni3c}, we adopt scene renders as camera conditions.

\noindent\textbf{Panoramic Representation.} A panorama image captures a full $360^\circ \times 180^\circ$ view of a scene from a fixed viewpoint, inherently representing visual signals on a spherical domain, as shown in Figure \ref{fig:per_pano}. This spherical nature is typically parameterized using spherical coordinates $(\phi, \theta)$, where $\phi$ denotes the azimuth angle and $\theta$ the elevation. 
An panorama video is a sequence of panorama images, which can represent a large-scale wide-coverage 3D scene.
Due to its wide field of view, panoramic data is commonly used in applications such as scene reconstruction, and immersive content generation.

\begin{figure}
    \centering
    \includegraphics[width=\linewidth]{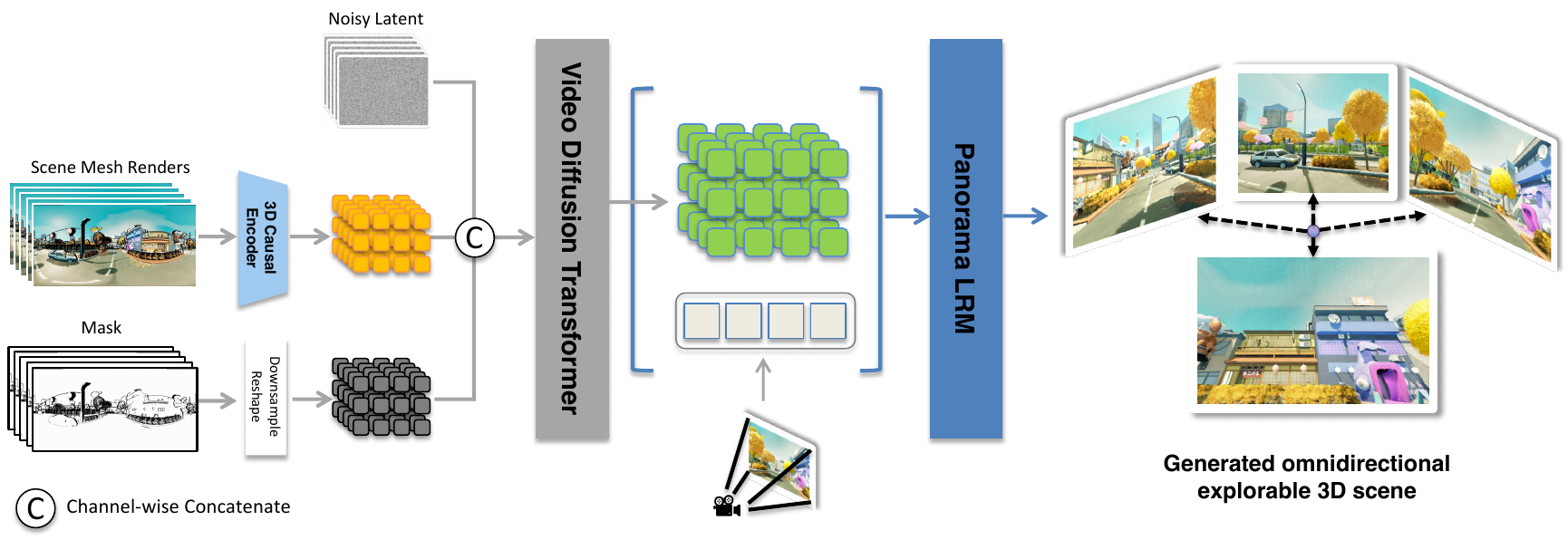}
    \caption{Core components of our framework. Given trajectory guidance in the form of scene mesh renderings and corresponding masks, obtained by rendering an estimated mesh along a user-defined camera trajectory, we train an image-to-video diffusion model to generate high-quality panoramic videos that precisely follow the specified trajectory. The generated 2D panoramic content is then lifted into an omnidirectional, explorable 3D world using a large-scale panorama reconstruction model.}
    \label{overview}
\end{figure}

\section{Method}

Our goal is to generate a geometric consistent omnidirectional explorable 3D world from a single image or text input. 
To achieve this, we leverage the panoramic representation and combine it with image diffusion and  video diffusion priors. 
The overall framework consists of three stages.  
We firstly utilize LoRA-based image diffusion models to convert input text or perspective image to a panorama image, whose depth can be predicted by methods like MoGe \cite{wang2025moge}. We leave details of panorama image generation in Appendix \ref{app:image}.  Next, as illustrated in Figure~\ref{overview}, a trajectory-constrained panorama video will be generated following the panorama image with its depth, as introduced in Section~\ref{sec:traj}. Finally, the generated panorama video will be lifted to an 3D world by two approaches, which are described in Section~\ref{sec:3d}.

\subsection{Trajectory Guided Panorama Video Generation}
\label{sec:traj}

\noindent \textbf{Trajectory Guidance Construction}. Given an input panorama image with depth and a predefined camera trajectory which reflects the range of generated scene, we aim to generate a panorama video whose camera movements exactly follow the input trajectory. We accomplish this by construct a sequence of scene renders and use them as trajectory guidance for a conditional panorama video generation model. To produce the trajectory guidance,  we firstly construct a polygonal mesh from the input panorama image with its depth, and then render it to a video following the input trajectory. 

Specifically, given a panorama $\mathbf{I}_0 \in \mathbb{R}^{H \times W \times 3}$ with its depth map $\mathbf{D}_0 \in \mathbb{R}^{H \times W \times 1}$, we construct the initial scene mesh $\mathbf{S}$ by projecting depth values into world coordinates to get mesh vertex positions. Then these vertex are connected to mesh faces according to their connectivity in the pixel space. The color of each mesh vertex is determined by its corresponding pixel color in $\mathbf{I}_0$. To represent invisible areas of $\mathbf{I}_0$ and accurately capture its occlusion relationship with visible parts, we further select pixels with drastically changing depth, mark their corresponding vertices as invisible and assign pure black color on them. In detail, the depth change of each pixel is calculated the as depth variation among its 1-ring neighbor pixels, and those pixels with depth change larger than a predefined threshold will be marked as invisible.

\begin{figure}[t]
    \centering
    \includegraphics[width=0.95\linewidth]{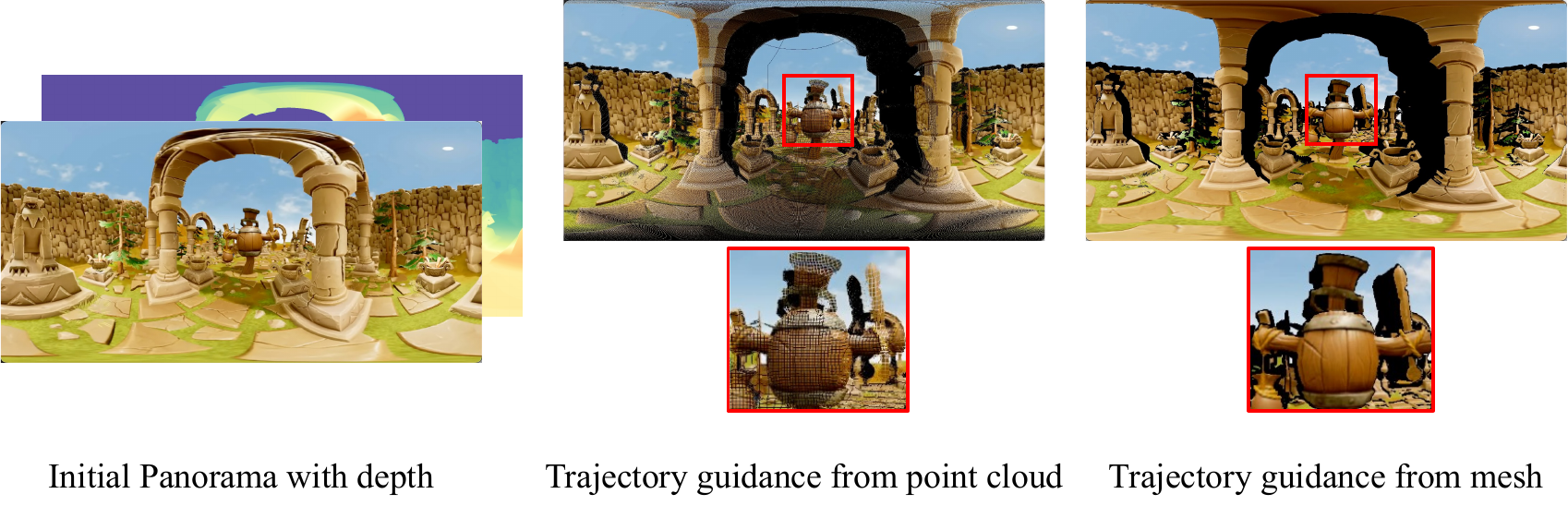}
    \caption{Comparison of trajectory guidance derived from mesh and point cloud representations. Results guided by point clouds suffer from noticeable artifacts, which degrade generation quality.}
    \label{mesh}
\end{figure}

With the initial scene mesh $\mathbf{S}$ and a predefined camera trajectory $\mathbf{C}$ which contains $N_f$ frames of camera poses, we can get a sequence of scene mesh renders $\{(\mathbf{I}_i, \mathbf{M}_i)\}_{i=1}^{N_f}$, which serves as trajectory guidance. Here,  $N_{f}$ represents the number of frames of $\mathbf{C}$, $\mathbf{I}_i$ and $\mathbf{M}_i$ represent the RGB and binary mask image rendered under the $i$-th camera pose. In the $i$-th frame, a pixel in $\mathbf{I}_i$ is marked as 1 in $\mathbf{M}_i$ if it is rendered from a visible face with all vertices visible; otherwise, the corresponding value in $\mathbf{M}_i$ is set to 0. As shown in Figure~\ref{mesh}, previous works~\cite{yu2024viewcrafter,yu2025trajectorycrafter} typically use point cloud renders as conditions, which often suffer from Moiré patterns and incorrect occlusion between objects. In contrast, our proposed scene mesh renders effectively reduce these geometric artifacts, leading to improved video generation quality.

\noindent\textbf{Panoramic Video Generation}. We adapt the image-to-video (I2V) diffusion model to generate panoramic videos. 
Given the trajectory-guided video sequence $\{(\mathbf{I}_i, \mathbf{M}_i)\}_{i=1}^{N_{f}}$, we first encode $\{\mathbf{I}_i\}$ using the video model's 3D causal VAE encoder to obtain video latents. The corresponding mask sequence $\{\mathbf{M}_i\}$ is downsampled and reshaped to mask latents that match the dimension of the video latents. Afterwards, both video and mask latents are concatenated with the noisy latent $\epsilon$ and then fed into the Diffusion Transformer backbone for denoising. To provide global semantic guidance, we also inject the input panorama $\mathbf{I}_0$ with its correlated or annotated text prompt into the network. Specifically, $\mathbf{I}_0$ is encoded using CLIP~\cite{radford2021learning} to obtain an image-level embedding, which is then fused with the text prompt embedding and integrated into the generation process via cross-attention.
We also introduce a LoRA module to enable stable and fast training. During training, the base model remains frozen while only the LoRA parameters are updated.

\subsection{3D World Generation}
\label{sec:3d}
To generate explorable consistent 3D world, we need to bridge the gap between 2D video generation and 3D scene reconstruction. 
We provide two methods: an optimization-based method for detailed 3D scene reconstruction, and a large panoramic reconstruction model that support rapid reconstruction.

\subsubsection{Optimization-based 3D Reconstruction}

The optimization-based method takes the generated video with the input camera trajectory as input. 
Due to the substantial redundancy present in adjacent frames, we construct a set of keyframes by selecting one panoramic frame every five frames, and use only these keyframes for optimization.
We notice that for the original 3DGS optimization pipeline \cite{kerbl20233d}, the quality of initialization significantly influences the optimized 3D scene. 
Therefore, we estimate the panorama depth of the keyframes with MoGe~\cite{wang2025moge}, perform the least squares registration to align these resulting depth maps, and use their corresponding world-coordinate point clouds as the initialization for 3DGS blobs. Since the original 3DGS optimization pipeline only accepts perspective images as input, we first crop each panoramic keyframe into 12 perspective images. We then apply StableSR~\cite{wang2024exploiting} to perform super-resolution on these perspective images, which are subsequently fed into the 3DGS optimization pipeline~\cite{kerbl20233d}.
The L1 loss between the rendered images and the input images is used as the objective function.

\subsubsection{Large Panorama Reconstruction Model}

Several recent works have explored feed-forward 3D scene reconstruction~\cite{liang2025wonderland, ziwen2024long} for efficient generation of 3D content. Inspired by Wonderland~\cite{liang2025wonderland}, we aim to directly infer 3D Gaussians~\cite{kerbl20233d} from generated video latents to reduce memory consumption caused by transforming raw images to a large number of tokens, leveraging their perceptual equivalence to images~\cite{rombach2022high}.

Given the video latent representation $\mathbf{z} \in \mathbb{R}^{t \times h \times w \times c}$ and the corresponding camera poses encoded as spherical Plücker embeddings $p \in \mathbb{R}^{T \times H \times W \times 6}$, we first transform them into latent tokens and pose tokens using dedicated patchify modules to ensure equal sequence lengths. These tokens are then concatenated along the channel dimension and fed into a series of base Transformer blocks. For Gaussian attribute prediction, we adopt the DPT head \cite{ranftl2021vision}.
Since the DPT head performs upsampling only along the spatial dimensions, we further employ a 3D deconvolution layer to upsample along the temporal dimension, ensuring alignment with the original video sequence on temporal dimension. Finally, the 3D deconvolution layer produces the 3D Gaussian attributes $\mathbf{G} \in \mathbb{R}^{T \times \frac{H}{n} \times \frac{W}{n} \times 12}$ as a 12-channel tensor, consisting of 3 channels for RGB color, 3 for scale, 4 for rotation (represented as a quaternion), 1 for opacity, and 1 for depth, where $n$ denotes the spatial downsampling factor. 
More details of our model architecture can be found in the Appendix \ref{app:lrm}.

\noindent \textbf{Optimization}.
Previous panorama 3DGS reconstruction methods \cite{lee2024odgs} have primarily focused on rendering panorama images from 3DGS and computing photometric loss in the panoramic image domain.
While a straightforward approach for optimizing our model would involve rendering panoramic images for loss computation, we empirically found that using panoramic images as direct supervision introduces sparsity artifacts when rendered with a perspective 3DGS rasterizer, significantly hindering practical application.
To address this, we introduce an alternative strategy: for each selected panoramic image, we first generate 12 perspective‐view patches that collectively cover the full $360^\circ$ field. Because using all 12 patches as supervision is computationally prohibitive, we randomly sample one patch per panorama and render it with a standard perspective rasterizer. To further guard against overfitting to observed viewpoints, we augment this reference set with both interpolated and extrapolated views. During training, we then compute our reconstruction loss between each rendered perspective patch and its corresponding ground‑truth crop, driving accurate panoramic 3D reconstruction.

\noindent \textbf{Two Stage Training Strategy}.
Due to the substantial domain gap between video latents and 3DGS, training the panorama reconstruction model poses significant challenges when estimating panoramic depth from video latent. Although the latent retains appearance consistency with the original images, it lacks explicit geometric cues, making depth prediction particularly difficult.

To overcome the challenge of panoramic depth estimation from video latent, we adopt a two stage training strategy. In the first stage, we initial  the
model training with depth loss. Specifically, our goal is to predict metric depth for each frame based on the video latent. 
Different from DepthPro \cite{bochkovskii2024depth} that adopts inverse depth as the training target, we focus on absolute depth. This approach emphasizes regions closer to the camera, which are typically more critical for accurate reconstruction than distant areas.
Moreover, to accelerate the two-stage optimization of other GS attributes, we incorporate a harmonic loss in the first stage to encourage the predicted 3-channel RGB colors to align with the corresponding pixel values.
The total training objective in the first stage is defined as:
\begin{equation}
\mathcal{L}_{\text{first\_stage}} = \mathcal{L}_{\text{Smooth-L1}}(\hat{\mathbf{D}}, \mathbf{D}) + \lambda_1 \mathcal{L}_{\text{L1}}(\hat{\mathbf{H}}, \mathbf{H}),
\end{equation}
where $\hat{\mathbf{D}}$ and $\mathbf{D}$ denote the predicted and ground-truth inverse depth maps, respectively. Moreover, $\hat{\mathbf{H}}$ and $\mathbf{H}$ represent the predicted and ground-truth harmonic RGB components, and $\lambda_1$ is the loss weight.

For the second stage, we optimize the remaining GS attributes using an image reconstruction loss that combines mean squared error (MSE) and perceptual similarity (LPIPS) loss \cite{zhang2018unreasonable}, defined as:
\begin{equation}
\mathcal{L}_{\text{second\_stage}} = \mathcal{L}_{\text{MSE}}(\hat{\mathbf{I}}, \mathbf{I}) + \lambda_2 \mathcal{L}_{\text{LPIPS}}(\hat{\mathbf{I}}, \mathbf{I}),
\end{equation}
where $\hat{\mathbf{I}}$ and $\mathbf{I}$ denote the rendered novel-view image and the corresponding ground-truth image, respectively, and $\lambda_2$ is a weighting factor. 

In the second stage, we freeze the depth‐prediction parameters and update only the remaining Gaussian attributes. To avoid overfitting to observed views and improve generalization to novel viewpoints, we randomly sample 32 reference views comprising three categories: views seen in the context frames, interpolated views, and extrapolated views, where interpolated views and extrapolated views are sampled novel views. Each panoramic view is then cropped into 12 perspective images that can include the whole panorama at 512 $\times$ 512 resolution, with a randomly selected field of view between $60^\circ$ and $120^\circ$. These crops serve as supervision signals during GS attribute fine‑tuning.

\begin{figure}[t]
    \centering
    \includegraphics[width=0.9\linewidth]{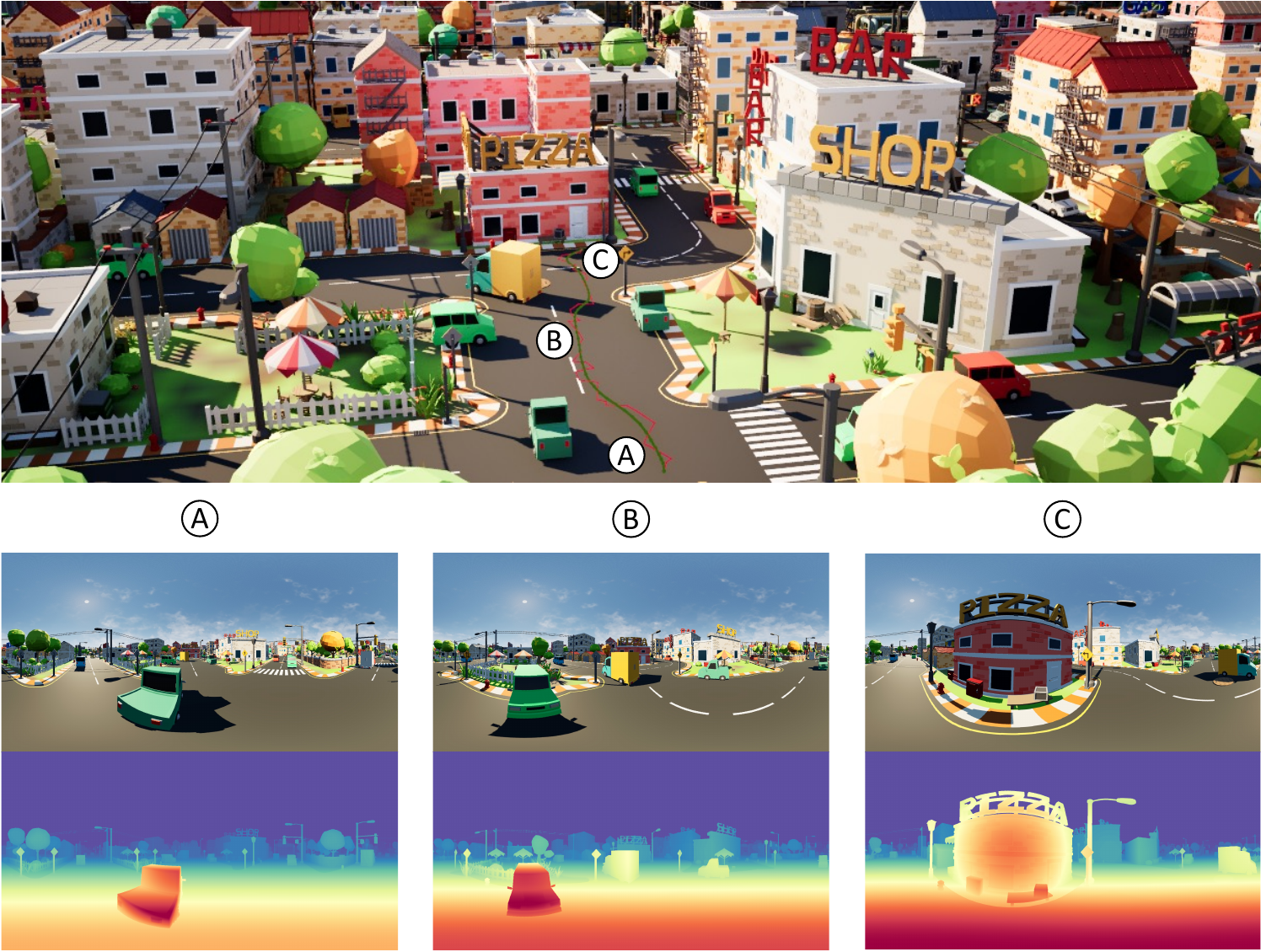}
    \caption{Dataset Illustration. We present a scene from the dataset and the data collection process. Two points are firstly randomly sampled on the route and connected by their shortest path as the red line segments shows. Then, a Laplacian smoothing algorithm is applied on the initial path to reach the smooth greed path, where paranoma videos and depths will be recorded along with their camera poses. We also show the captured rgb image and depth map of three different frames on the bottom of the figure.}
    \label{fig:dataset}
\end{figure}

\section{Matrix-Pano Dataset}

Training our proposed models requires a panorama dataset with detailed camera trejectory annotation and corresponding depth maps, which are not provided in any existing dataset. 
To tackle this issue, we present the Matrix-Pano Dataset, a scalable synthetic dataset designed for generating omnidirectional explorable 3D worlds. Created entirely within Unreal Engine 5, it offers high-fidelity simulations, diverse environments, and precise annotation over camera trajectories. The construction process of our dataset can be viewed in Figure \ref{fig:dataset}, which consists of the following steps.

\textit{Step 1: Data Collection in Unreal Engine.}
Unreal Engine allows for flexible data preparation across a variety of scenes, perspectives, and content. Utilizing 504 high-quality 3D scenes, we encompass a wide range of indoor and outdoor settings with varying weather and lighting conditions, forming the foundation for trajectory-based video recording.

\textit{Step 2: Exploration Route Sampling.}
We developed a trajectory sampling algorithm to generate plausible and visually coherent camera paths. For each scene, we first identify walkable surfaces, e.g., roads or floors, then apply the Delaunay triangulation algorithm, which creates a set of non-overlapping triangular meshes from sparse points on a two-dimensional plane. As shown in Figure~\ref{fig:dataset}, path candidates are then sampled through three steps: (1) randomly selecting two mesh vertices, (2) calculating the shortest path using Dijkstra's algorithm (the red lines), and (3) applying Laplacian smoothing to minimize abrupt turns (the green curves). Only trajectories longer than 18 meters are kept to ensure adequate temporal dynamics.

\begin{table}[t]
\centering
\scriptsize
\caption{Comparison of panorama datasets for 3D world generation.}
\begin{tabular}{lccccccc}
\toprule
\textbf{Dataset} & \textbf{\#Samples} & \textbf{Diversity} & \textbf{Cam. Pose} & \textbf{Depth} & \textbf{Text} & \textbf{Resolution} & \textbf{Video} \\
\midrule
Imagine360~\cite{tan2024imagine360} & 10k   & Low & \ding{55} & \ding{55} & \ding{51} & Mixed   & \ding{51} \\
Web360~\cite{wang2024360dvd}     &  2k   & Low & \ding{55} & \ding{55}   & \ding{51} & 720×1028 & \ding{51} \\
Argus \cite{luo2025beyond} & 283k   & Medium & \ding{55} & \ding{55}  & \ding{55} &Mixed  & \ding{51} \\
360-1M \cite{wallingford2024image} & 1,076k   & High & \ding{51} & \ding{55}  & \ding{55} &Mixed  & \ding{51} \\
PanoWan~\cite{xia2025panowan}    & 13k  & High   & \ding{55} & \ding{55}   & \ding{51} & Mixed  & \ding{51} \\
Matrix-Pano (Ours)        & \textbf{116k} & \textbf{High} & \ding{51} & \ding{51}  & \ding{51} & 1024×2048 & \ding{51} \\
\bottomrule
\end{tabular}
\vspace{0.1cm}
\label{tab:dataset_comparison}
\end{table}

\textit{Step 3: Collision Detection.}
We implement a collision detection mechanism to remove trajectories that cause geometry clipping or object intersections, which can degrade quality and stability. Using a bounding box proxy algorithm, objects are simplified into 3D bounding boxes based on their nearest and farthest points, balancing spatial accuracy with computational efficiency. Trajectories are simulated step-by-step, and any intersecting paths are discarded.

\textit{Step 4: Data Annotation and Quality Filtering.} We ensure the quality of the dataset through two filtering stages:

(1) Automatic Filtering: We use Video-LLaMA3 \cite{damonlpsg2025videollama3} to evaluate videos based on detailed quality, semantic information, and motion richness, filtering out low-quality content.

(2) Manual Assessment: The first frame of each video is manually reviewed to remove samples with poor rendering quality or missing details. Finally, Video-LLaMA3 automatically annotates videos to support text-controlled and multimodal tasks.

Following this multi-step process, we retain 116K high-quality static panoramic video sequences, each paired with its corresponding 3D exploration route and depth map. More details about data preparation can be found in Appendix~\ref{app:dataset}. We compare our dataset with existing panoramic datasets in terms of the number of samples, scene diversity, availability of camera poses and depth, support for long-range exploration, presence of text annotations, resolution, and whether the dataset includes video sequences in Table \ref{tab:dataset_comparison}. Our dataset uniquely provides comprehensive and accurate information on both camera parameters and depth maps.

\section{Results}

\subsection{Dataset and Evaluation Metric}
\noindent\textbf{Dataset.}
 From the proposed Matrix-Pano dataset, we crop and extract 200K video clips from all 116K video sequences, each containing 81 frames, to train our panoramic video generation model. For each video clips, we construct the scene mesh render and the panorama mask with its ground truth depth. 
For evaluation, we select 200 panoramic videos from our synthetic dataset as the test set, ensuring no overlap with the training data.

\noindent\textbf{Evaluation Metric.}
To evaluate the visual quality and temporal coherence of generated videos, we adopt Fréchet Inception Distance (FID) \cite{heusel2017gans} and Fréchet Video Distance (FVD)\cite{unterthiner2019fvd} as our  metrics.
To assess camera controllability, we follow WorldScore \cite{duan2025worldscore} and compute the rotation error ($R_\text{err}$) and translation error ($T_\text{err}$). We use VGGT~\cite{wang2025vggt} to accurately estimate camera poses for all methods. Moreover, we also compare PSNR, SSIM, and LPIPS between the generated videos with ground-truth video clips.

\subsection{Implementation Details}

We build our trajectory-constrained panoramic video generation model upon Wan2.1-I2V-14B\cite{wan2025wan}. We train two models at resolution of $480 \times 960$ (480p) and $720 \times 1440$ (720p) respectively. During training, each video contains 81 frames. The network is trained for 6000 iterations on a curated dataset of 200K panoramic video sequences, using a learning rate of $1 \times 10^{-4}$ and batch size of 21. 
For the large panoramic 3D reconstruction model, our network largely follows the architecture of Wonderland \cite{liang2025wonderland}, with the key modification of adopting a DPT head \cite{ranftl2021vision} to predict 3D GS attributes and sphere projection for 3D GS mean prediction. The large reconstruction model is trained on the basis of the 480p panorama video generation model.
We selected 81 frames from a video sequence as context inputs with a random stride between 1 and 3. 
For supervision, we adopt the sampling strategy proposed in Wonderland \cite{liang2025wonderland}, where a total of 32 reference views are selected per iteration from three types of frames: context frames, interpolated frames, and extrapolated frames.
And the model is trained with a learning rate of $1 \times 10^{-4}$.
During the inference stage, the initial depth map for the input panorama can either be supplied by the user or estimated using depth-estimation methods such as MoGe~\cite{wang2025moge}.

\subsection{Comparison with state-of-the-art methods}
\begin{figure}[t]
    \centering
    \includegraphics[width=\linewidth]{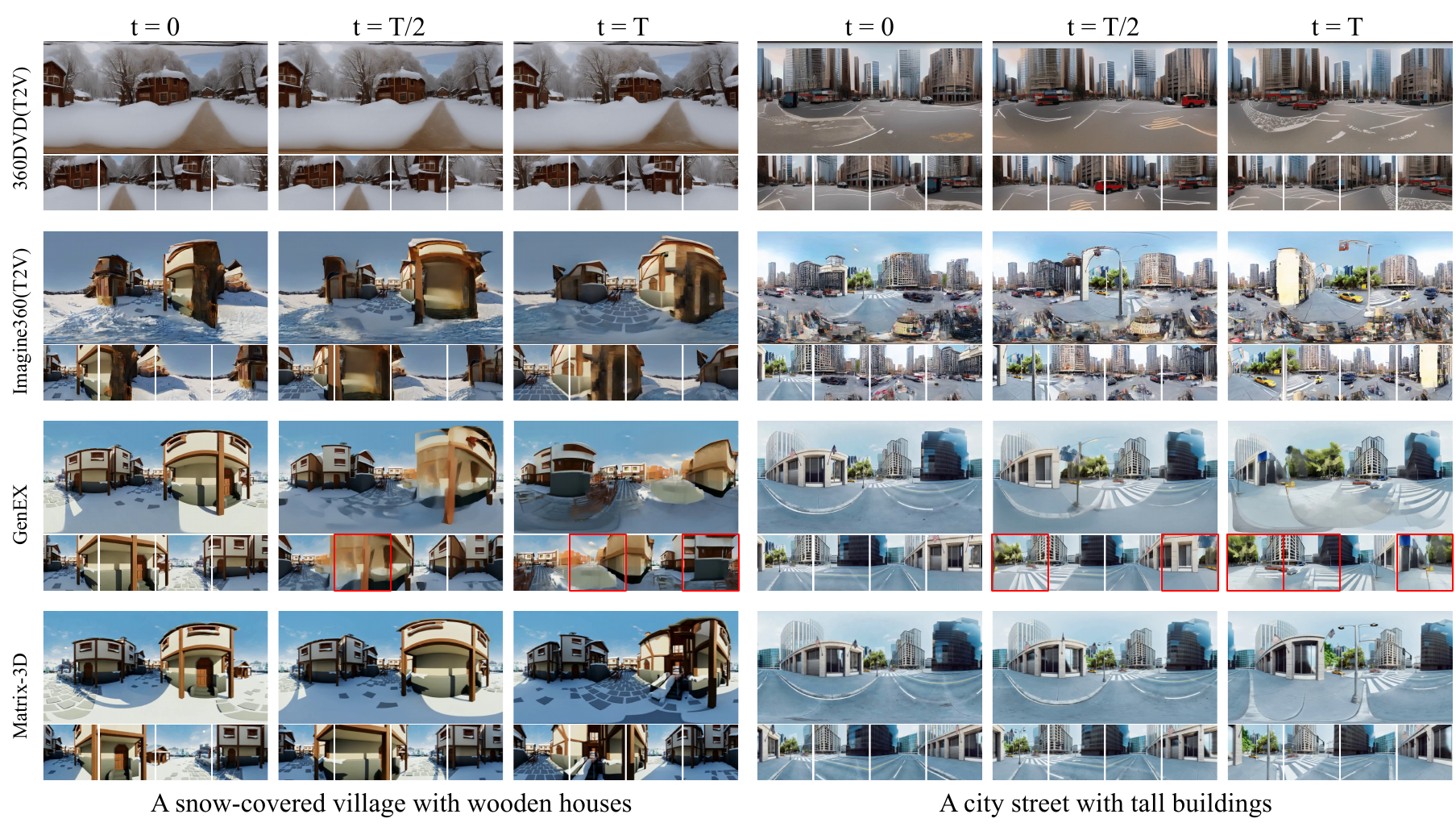}
    \caption{Qualitative comparison of panorama video generation methods. For each generated video, we extract panoramic images from the first frame (\(t = 0\)), the middle frame (\(t = T/2\)), and the last frame (\(t = T\)), along with their four orthogonal perspective views listed below. Matrix-3D produces panoramic videos with exceptional visual quality and geometric consistency.
}
    \label{fig:pc_panocomp}
\end{figure}

\noindent\textbf{Comparison with Panorama Video Generation Model. }
We evaluate the effectiveness of our proposed method by comparing it against state-of-the-art panoramic video generation models. Specifically, we select the following three representative baselines: (1) 360DVD 
\cite{wang2024360dvd}, which incorporates a trainable 360-Adapter to adapt standard text-to-video (T2V) models to the panoramic domain; (2) Imagine360 \cite{tan2024imagine360}, a model designed to convert perspective videos into panoramic ones. For evaluation, we crop perspective videos from ground-truth panoramas and assess the model's generation performance; (3) GenEx \cite{lu2024genex}, a panoramic video generation model built upon stable video diffusion \cite{blattmann2023stable}. Since the official release of GenEx only supports image-to-video generation and lacks a controllable video generation checkpoint, we compare its video quality with our method but exclude it from controllability-related evaluations. We report the quantitative comparison in Table \ref{tab:combined} and show the visual differences in Figure \ref{fig:pc_panocomp}. The generated results of 360DVD, Imagine360 and GenEx exhibit more distorted patterns and a blurrier visual appearance in both scenarios. In contrast, Matrix-3D delivers superior visual quality and plausible geometric structure in the generated panorama videos. 

\begin{table}[t]
    \centering
    \small
    \caption{Comparison of Panoramic Video Generation  and Camera Guided Generation Models.}
    \label{tab:combined}
    \setlength{\tabcolsep}{5pt}
    \begin{tabular}{lcccccccc}
        \toprule
        {Models} & {PSNR} $\uparrow$ & {SSIM} $\uparrow$ & {LPIPS} $\downarrow$ & {FID} $\downarrow$ & {FVD} $\downarrow$ & {$R_{err}$} $\downarrow$  & {$T_{err}$} $\downarrow$ \\
        \midrule
        \multicolumn{8}{l}{\textit{Panoramic Video Generation Models}} \\
        \midrule
360DVD~\cite{wang2024360dvd}           &9.65	&0.349	&0.834 	&112 	&2700 & -- & -- \\
Imagine360~\cite{tan2024imagine360}    &11.6	&0.391	&0.599 	&66.7	&1600 & -- & -- \\
GenEx~\cite{lu2024genex}               &16.1	&0.600	&0.380 	&42.2	&1110 & -- & -- \\
Matrix-3D 480p                              &23.7	&0.722	&\textbf{0.0776}	&15.4	&234 & -- & -- \\
Matrix-3D 720p                               &\textbf{23.9}	&\textbf{0.747}	&0.0907	&\textbf{11.3}	&\textbf{140} & -- & -- \\
        \midrule
        \multicolumn{8}{l}{\textit{Camera Guided Generation Models}} \\
        \midrule
ViewCrafter \cite{yu2024viewcrafter}             &21.6	&0.701	&0.161	&47.3	&762	&0.0940 	&0.0453 \\
TrajectoryCrafter \cite{yu2025trajectorycrafter}       &21.8	&0.682	&0.126	&33.1	&675	&0.0338	        &0.0488 \\
Matrix-3D 480p Persp.          &24.1	&0.750	&0.113	&23.9	&438	&0.0325	        &0.0310 \\
Matrix-3D 720p Persp.          &\textbf{24.3}	&\textbf{0.777}	&\textbf{0.108}	&\textbf{12.5}	&\textbf{165}	&\textbf{0.0306}	        &\textbf{0.0297} \\
        \bottomrule
    \end{tabular}
    \vspace{-3mm}
\end{table}

\noindent\textbf{Comparison with Camera-Guided Video Generation Model. }
To the best of our knowledge, we are the first to address controllable panoramic video generation. As no existing work shares the same task formulation, direct comparisons are not feasible. Instead, we evaluate our method against state-of-the-art camera-controllable perspective video generation models. 
We select two representative methods for this evaluation including  ViewCrafter \cite{yu2024viewcrafter} and TrajectoryCrafter~\cite{yu2025trajectorycrafter}. For a fair comparison, the input geometry of ViewCrafter, TrajectoryCrafter and our method are all initialized by the annotated depth of the evaluation dataset. Since only Matrix-3D generates panoramic videos, we crop our results into perspective view videos with a 90-degree field of view (FoV) before conducting the evaluation.
We assess the generation performance from two key aspects including  image quality and camera controllability. Our method outperforms previous approaches regarding visual quality and camera controllability, as evidenced by the quantitative results in Table \ref{tab:combined}. Our approach demonstrates a substantial improvement in camera controllability, enabling more accurate and flexible camera guidance.

\begin{figure}[t]
    \centering
    \includegraphics[width=\linewidth]{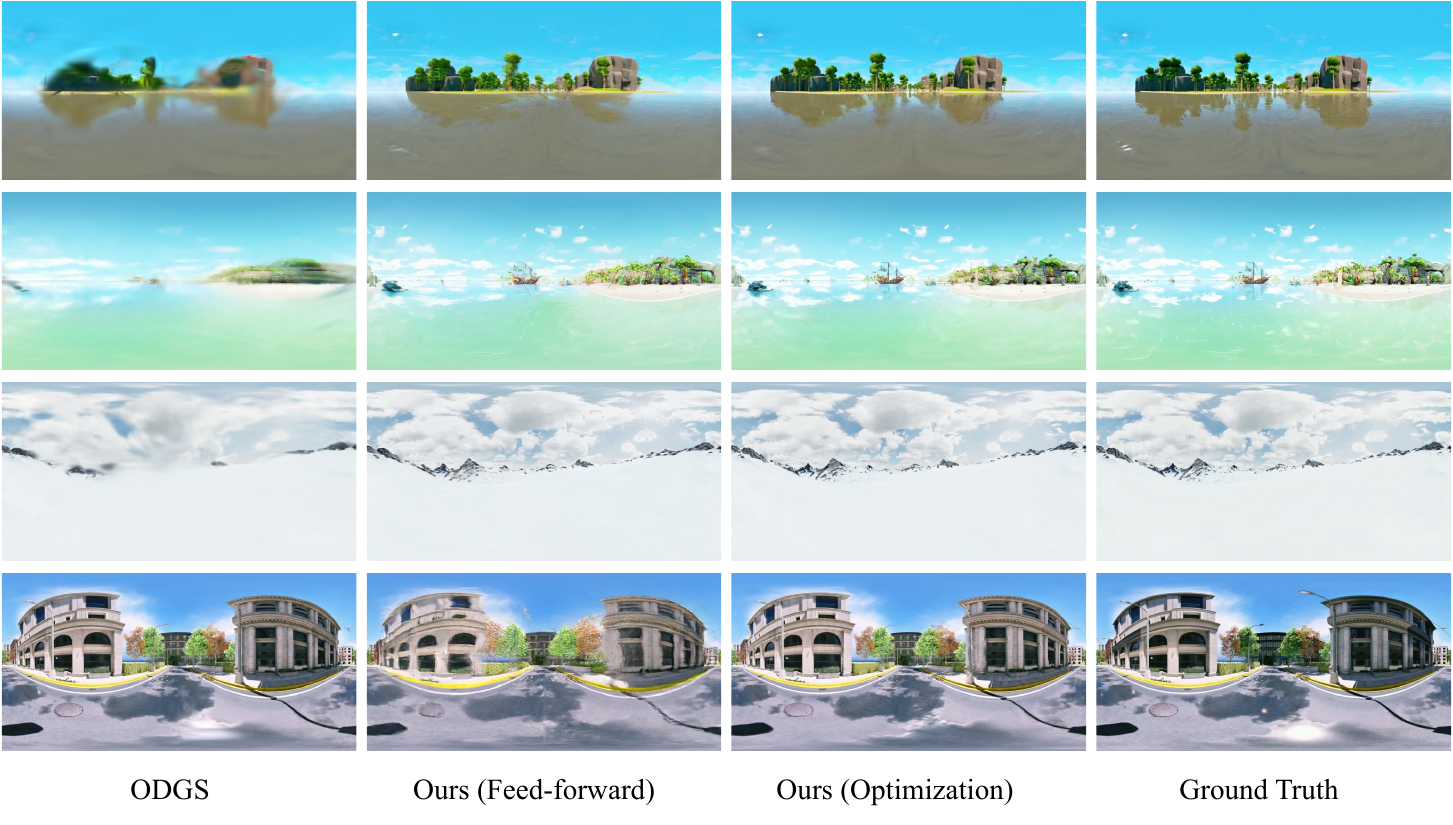}
    \caption{Qualitative comparison of 3D world reconstruction.}
    \label{fig:3d_comp}
\end{figure}

\noindent\textbf{Comparison with 3D World Reconstruction. }
To evaluate the performance of 3D world reconstruction, we compare our methods which includes optimization-based method and feed-forward based method against ODGS~\cite{lee2024odgs}, a state-of-the-art optimization-based 3DGS approach tailored for panoramic inputs.
Our proposed optimization-based pipeline substantially outperforms ODGS in reconstruction quality, while our feed-forward variant enables fast and efficient reconstruction. We visualize the reference image alongside the rendered panoramic image in Figure \ref{fig:3d_comp} and report the quantitative comparison in Table \ref{tab:3d_quantitative}.

\begin{table}[t]
\centering
\small
\caption{Quantitative comparison of 3D world reconstruction on the benchmark dataset. All metrics except for time are calculated on perspective images cropped from the panorama renderings. Our optimization-based pipeline achieves the best performance in terms of visual quality, while the feed-forward pipeline enables much faster reconstruction.}
\label{tab:3d_quantitative}
\begin{tabular}{lcccc}
\toprule
Methods &PSNR ↑ & LPIPS ↓ & SSIM ↑   & Time (↓) [s] \\
\midrule
ODGS~\cite{lee2024odgs} &22.04 &0.444  & 0.673 & 745 \\
Ours (Feed-forward) &22.30 &0.389  &0.647  &\textbf{10}  \\
Ours (Optimization-based) &\textbf{27.62} &\textbf{0.294}  &\textbf{0.816}  &571  \\
\bottomrule
\end{tabular}
\end{table}

\begin{figure}
    \centering
    \includegraphics[width=0.8\linewidth]{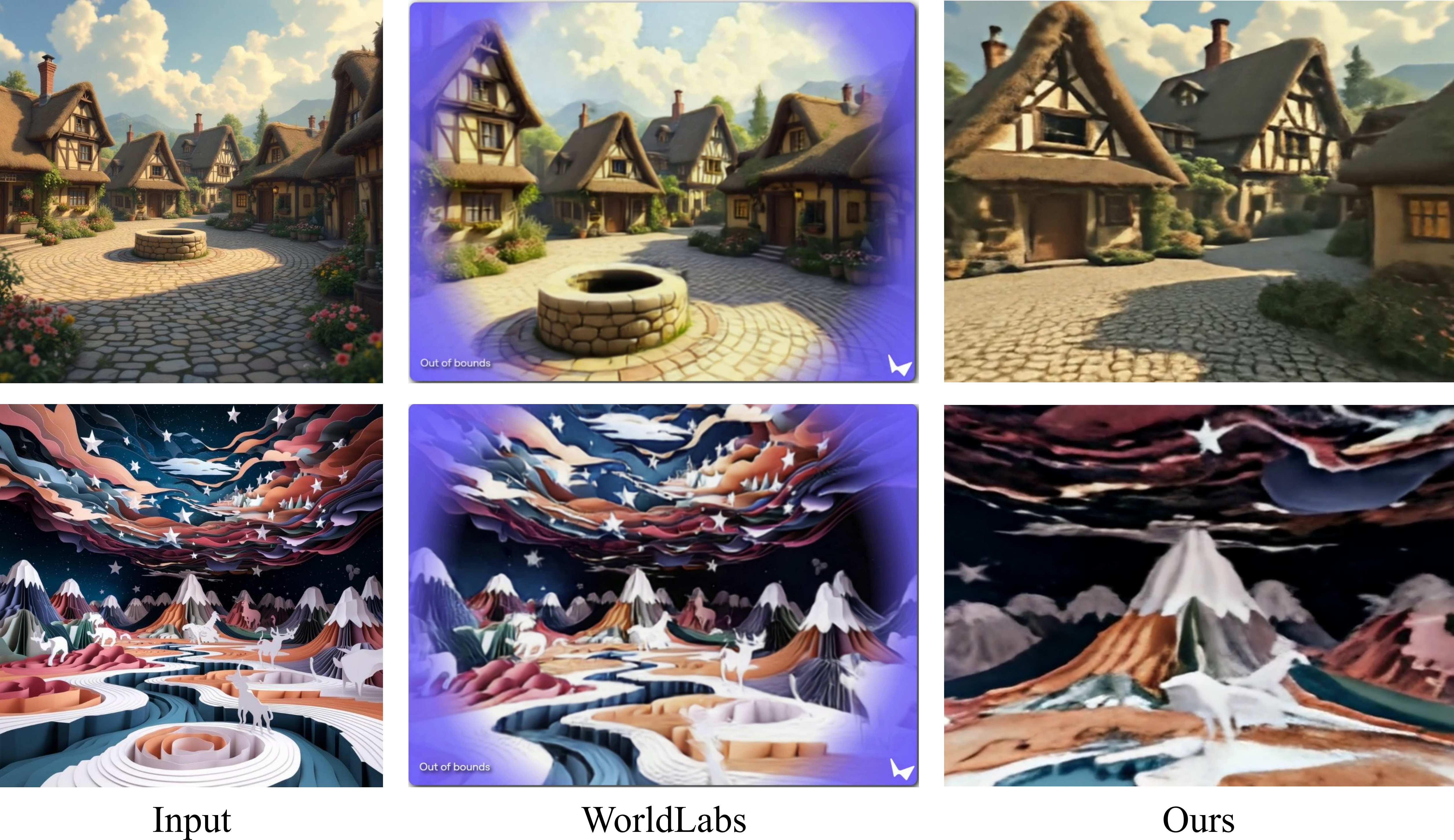}
    \caption{Comparison with WorldLabs~\cite{worldlabs2025}. Both methods generate 3D scenes from the same input image, and we show  the renderings of the 3D scene at the furthest reached positions, all under the same field of view. Matrix-3D can generate 3D scenes with greater range than WorldLabs.
}
    \label{fig:worldlabs}
\end{figure}

\begin{figure}
    \centering
    \includegraphics[width=0.95\linewidth]{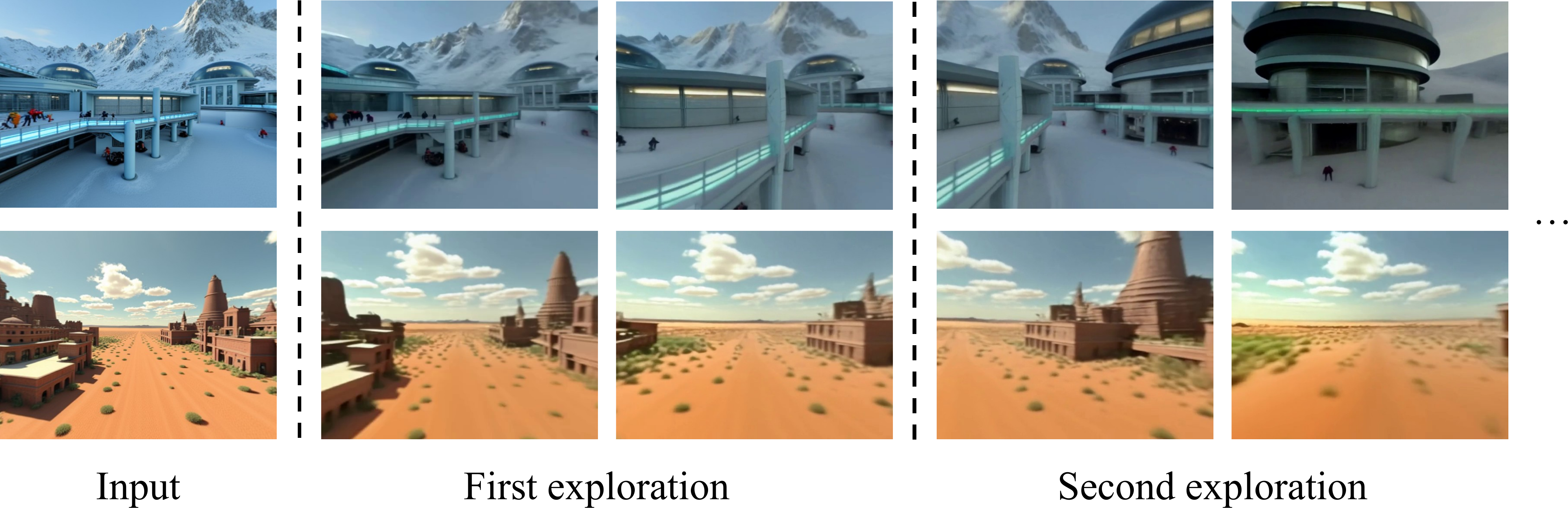}
    \caption{Endless Exploration. With an input image and initial trajectory, users can generate the first 3D scene segment and then explore further by changing direction and following a second trajectory. For ease of viewing, we present the perspective projection results from the center of each panorama.
}
    \label{fig:continue}
\end{figure}

\noindent\textbf{Comparison with  and WorldLabs.}
WorldLabs~\cite{worldlabs2025} releases its demo for generating 3D scenes from images and text in 2025. Compared to WonderWorld~\cite{yu2025wonderworld}, the results generated by WorldLabs exhibit more loop consistency. We conduct a simple comparison between WorldLabs and our method.  We generate scenes using both WorldLabs and our method from the same input image, then proceed linearly through the scene to evaluate the extent of navigable distance. Figure \ref{fig:worldlabs} shows the renderings of 3D scene at the furthest reached positions under the same FOV, demonstrating that the range of scenes generated by our method is significantly greater than that of WorldLabs. Please visit our \href{https://matrix-3d.github.io/}{website} for a more intuitive comparison.

\noindent\textbf{Endless Exploration.} 3D worlds generated by Matrix-3D allow exploration in any direction, facilitating the development of an endless exploration strategy, as shown in Figure \ref{fig:continue}. Given an input image and an initial trajectory path, users can generate the first segment of the 3D scene. Subsequently, users can look around, change direction, and continue exploration along a second trajectory. This approach enables endless exploration, allowing users to freely navigate the 3D scene in any direction. Please visit our \href{https://matrix-3d.github.io/}{website} for more details.

\begin{figure}[t]
    \centering
    \includegraphics[width=\linewidth]{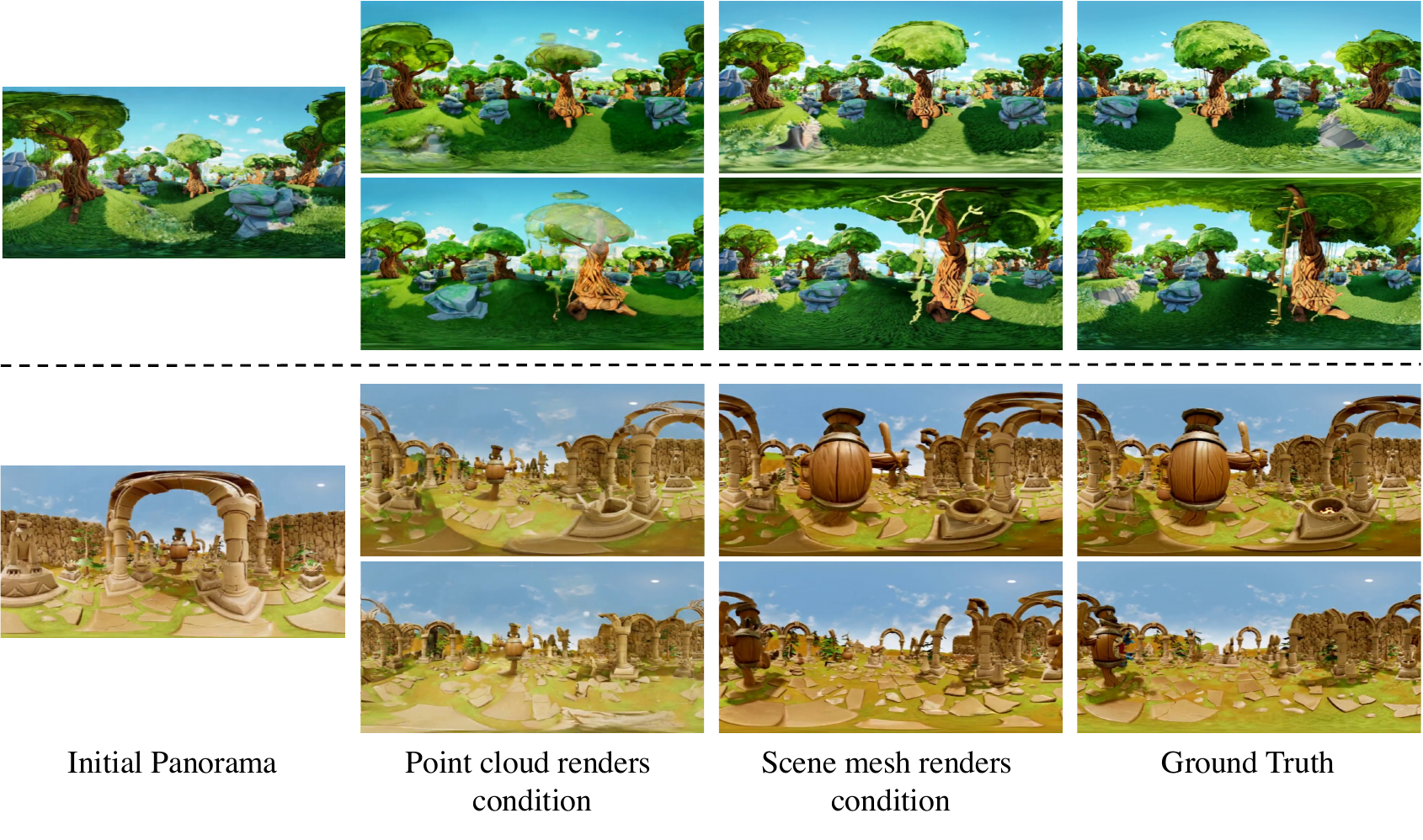}
    \caption{Qualitative comparison between trajectory guidance from point cloud and mesh renderings. Our method, which utilizes scene mesh renderings, significantly outperforms point cloud-based guidance in both geometric and textural consistency, camera controllability and visual quality.
}
    \label{fig:pc_ablation}
\end{figure}

\begin{table}[h]
    \centering
    \small
    \caption{Ablation study on different input conditions. Here $R_{err}$ and $T_{err}$ are calculated are calculated from the perspective video with a 90-degree FoV, cropped from the corresponding panoramic video.}
    \label{tab:pc_mesh}
    \setlength{\tabcolsep}{5pt}
    \begin{tabular}{lcccccccc}
        \toprule
        {Methods} & {PSNR} $\uparrow$ & {SSIM} $\uparrow$ & {LPIPS} $\downarrow$ & {FID} $\downarrow$ & {FVD} $\downarrow$ & { $R_{err}$} $\downarrow$  & {$T_{err}$} $\downarrow$ \\
        \midrule
point cloud renders  &23.4	&0.746	&0.0834	&15.9	&260	&0.0410	&0.0375 \\
mesh renders  &\textbf{23.8}	&\textbf{0.752}	&\textbf{0.0793}	&\textbf{15.3}	&\textbf{242}	&\textbf{0.0359}	&\textbf{0.0344} \\
    \bottomrule
    \end{tabular}
    \vspace{-3mm}
\end{table}

\subsection{Ablation Study}
\noindent\textbf{The effectiveness of rendered video from mesh.}
We conduct an ablation study to assess the effectiveness of using rendered videos from meshes instead of point clouds. Specifically, we train our panoramic video generation model on a subset of 5k data samples of 480p resolution under identical settings, varying only the source of trajectory guidance, i.e., using point cloud renders or scene mesh renders. Qualitative and quantitative comparisons are shown in Figure~\ref{fig:pc_ablation} and Table~\ref{tab:pc_mesh} respectively.
The visual quality, camera controllability, and geometric consistency of videos rendered from meshes significantly outperform those rendered from point clouds. 

\begin{figure}[ht]
  \centering
\includegraphics[width=\textwidth]{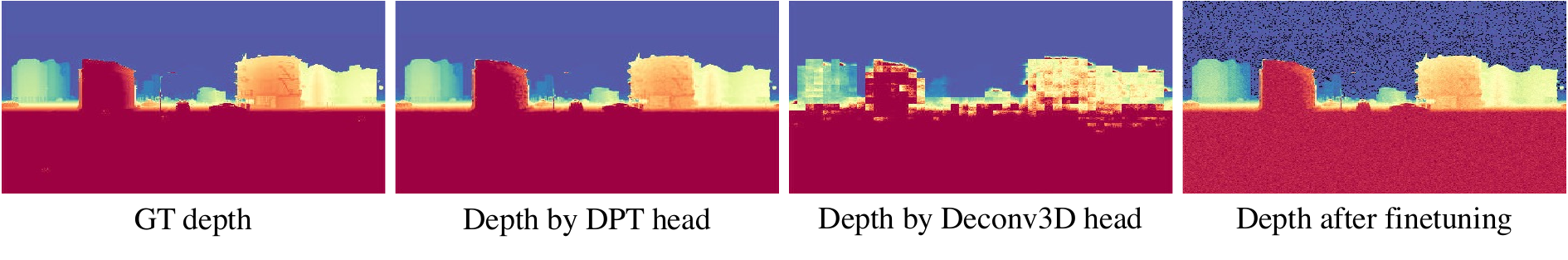}
    \caption{Comparison between depth predictions from the
    DPT head and the 3D deconvolution head. When depth-related parameters are not frozen, depth  degrades during second-stage training.}
  \label{fig:depth_ablation}
\end{figure}

\noindent\textbf{The effectiveness of DPT head.}
We compare depth prediction performance between a DPT head and a simple 3D deconvolution-based upsampling module. As shown in Figure~\ref{fig:depth_ablation}, the DPT head yields significantly more accurate depth estimations, demonstrating its superiority in capturing fine-grained spatial details by utilizing multi-scale information.

\noindent\textbf{The effectiveness of two stage training strategy.}
We conducted an ablation study to evaluate the effectiveness of our two-stage training strategy. We found that jointly predicting ray distances and other 3DGS attributes in a single stage fails to converge reliably, often resulting in unstable optimization and degraded reconstruction quality. Even when initializing the second stage from the first-stage checkpoint without freezing depth-related parameters, depth prediction degrades progressively during training. 
We show the qualitative results of fine-tuning without freezing depth-related parameters in Figure~\ref{fig:depth_ablation}. The depth prediction noticeably degrades, indicating the importance of preserving the pre-trained depth estimation capacity during fine-tuning.

\section{Conclusions and Future Work}
\noindent\textbf{Conclusions}
In this work, we present Matrix-3D, a unified framework for omnidirectional explorable and geometrically consistent 3D world generation from a single image or text prompt. 
By leveraging panoramic representations, our method enables high-quality and wide-coverage $360^\circ$ explorable scene generation. 
We first introduce a trajectory-guided panoramic video diffusion model conditioned on mesh renders, which produces visually coherent and structurally faithful scene videos.
To further lift the generated videos into 3D, we design two separate reconstruction pipelines: a feed-forward model for fast scene reconstruction and an optimization-based approach for high-fidelity geometry recovery. 
Additionally, we contribute the Matrix-Pano dataset — the first large-scale collection of synthetic panoramic videos with camera pose and depth annotations — to support training and evaluation. Extensive experiments validate the effectiveness of our approach, achieving state-of-the-art performance in both panoramic video generation and 3D world reconstruction. 
We believe our work paves the way for more robust and generalizable 3D world generation and spatial intelligence.

\noindent\textbf{Limitation}
Despite the advantages of our proposed framework, several limitations remain. Firstly, since our 3D scene generation model is built upon a video diffusion model, its inference speed is relatively slow, with the generation of a single scene taking tens of minutes.
Secondly, in the Matrix-Pano dataset, unnatural transitions in depth values occasionally occur in semi-transparent or perforated regions, such as those found in trees and fences.
Lastly, estimating depth from the video latent representation is particularly challenging: the latent compresses the original video and encodes only appearance cues, since the video VAE’s objectives do not incorporate geometric information. This imperfect depth estimation undermines the second-stage LRM training and, ultimately, degrades the final rendering quality.

\noindent\textbf{Future Work}
While our proposed Matrix-3D framework demonstrates strong performance in panoramic video generation and 3D world reconstruction, several promising directions remain for future exploration. 
Firstly, 3D scenes generated by Matrix-3D typically include information only from currently visible areas. Future research should investigate methods to generate scenes for unseen areas, such as by employing specific trajectory settings or integrating 3D object generation into the existing pipeline.
Secondly, the editability of the generated 3D worlds can be further enhanced. This involves enabling user-driven operations such as scene modification and semantic-level interactions, such as issuing high-level commands like ``add a tree beside the house'' or ``remove the car from the road''. 
Enhancing editability would make the system more adaptable for downstream applications in digital content creation, simulation environments, and embodied AI systems.
Finally, we aim to extend our method to dynamic scene generation, enabling each object in the scene to move and interact, thereby providing users with a more immersive experience and further advancing research in world models.

\bibliography{ref}
\bibliographystyle{plain}

\appendix

\clearpage
\section{Panorama Image Generation}\label{app:image}
For text-to-panorama generation, we train a LoRA on FLUX~\citep{flux2024} with 1000 selected panorama images from the Matrix-Pano dataset. Our image-to-panorama generation pipeline builds directly upon WorldGen~\citep{worldgen2025}. During the inference stage, we apply latent rotation and circular padding to enable the generation of loop-consistent results following PanFusion~\citep{panfusion2024}.

\section{LRM Network Architecture}\label{app:lrm}

\begin{figure}
    \centering
    \includegraphics[width=1\linewidth]{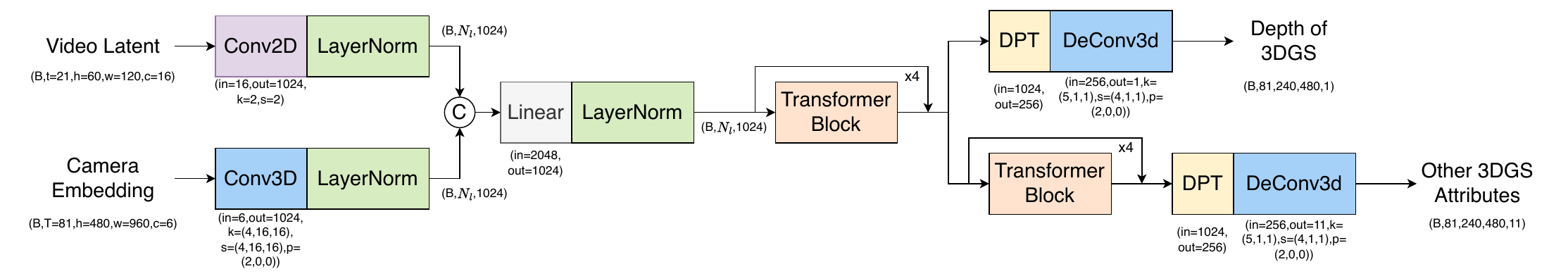}
    \caption{The network architecture of our large panorama reconstruction model.}
    \label{fig:LRM}
\end{figure}

As shown in Fig.~\ref{fig:LRM}, our large panorama reconstruction model takes video latents and camera embeddings as input to generate 3DGS representations. The video latent, generated from a video diffusion model or encoded from ground-truth video, is first processed by a 2D convolution layer followed by LayerNorm. Simultaneously, camera pose sequences are converted to spatiotemporal embeddings using a 3D convolution with kernel size $(4,16,16)$ and stride $(4,16,16)$, which aligns with the patchified structure of the video latent. The two token sequences are then concatenated along the channel dimension, linearly projected, and normalized.

The merged token features are fed into a sequence of 4 transformer blocks to capture global scene context. The resulting features are then decoded by two separate branches: one for estimating the depth of the 3D Gaussians and the other for predicting the remaining 3DGS attributes (e.g., rotation, color, opacity, and scale). Each decoding branch consists of  4-layer transformer module, followed by a depth-wise prediction transformer (DPT) and a 3D transposed convolution (DeConv3D) with upsampling strides $(r_t, 1, 1)$ and kernel size $(5,1,1)$.

The final outputs are: a depth volume of size $(B, 81, 240, 480, 1)$, and an attribute volume of size $(B, 81, 240, 480 11)$, which together form the complete 3DGS representation. During the second stage training, we frozen depth related parameters including the first 4-layer Transformer blocks.

\section{Dataset Construction Details}\label{app:dataset}
\subsection{Technical Challenges}
Achieving high-quality offline panorama video rendering in Unreal Engine, particularly for post-production stitching and virtual reality scenarios, involves several significant technical challenges.
\paragraph{Multi-directional Rendering and Precise Stitching}Panorama videos require offline rendering of six directions (front, back, left, right, up, down) for seamless stitching. For each frame, it is essential to ensure that the acquisition of all six views and the camera transform parameters are strictly synchronized. Any misalignment can easily lead to visual artifacts or dynamic ghosting.

\paragraph{Adaptation of Offline High-Quality Rendering Pipeline}While the Movie Render Queue (MRQ) in Unreal Engine supports offline high-fidelity rendering, its native workflow is not designed for multi-directional panorama capture and stitching. Automating the acquisition of six views per frame and guaranteeing absolute synchronization between rendered images and camera motion data within the MRQ framework presents a considerable technical challenge.

\paragraph{Interference from Screen Space Effects}By default, Unreal Engine enables various screen space post-processing effects such as Local Exposure, Vignette, and Bloom. These can introduce uncontrollable discrepancies in tone, brightness, and contrast between the six rendered views, severely compromising seamless panorama stitching.

\subsection{Proposed Solution}

To address these challenges, we propose a custom panorama rendering pipeline built on Movie Render Queue, with the following key technical innovations.

\paragraph{Custom Pass and Camera Synchronization System}
Custom passes are injected into the MRQ pipeline to automate offline rendering of all six directions (+X, -X, +Y, -Y, +Z, -Z) for each frame. The system ensures that the camera's transform (position and orientation) is locked and consistent across all directions per frame, thoroughly eliminating dynamic misalignment and tearing.

\paragraph{Strict Post-processing Exclusion and Exposure Locking}
All non-linear screen space post-processing effects—including dynamic exposure (Local Exposure), vignette, and tone mapping—are comprehensively disabled, ensuring physically consistent image properties across all directions. Exposure compensation is fixed at a constant value of 10.5 to prevent any automatic exposure fluctuation, resulting in an output with a controllable linear color space.

\paragraph{Intelligent Adaptation of Lumen Global Illumination}
Lumen is used as the global illumination solution, with parameters specifically tuned for panorama stitching. By restricting Lumen’s screen space sampling, we guarantee spatial consistency of global illumination across all six rendered views. At the same time, the rich indirect lighting and volumetric illumination characteristics of Lumen are preserved, achieving both physical seamlessness and high aesthetic quality.

\subsection{Technical Innovation and Value}
This pipeline systematically addresses the key challenges of offline panorama video production in Unreal Engine, enabling professional-grade, high-fidelity, seamless, and controllable panorama content creation. The proposed solution establishes a robust technical foundation for large-scale, high-quality content production in VR, film, and digital twin applications, as well as AI 3D scene training data, all based on Unreal Engine.

\end{document}